\documentclass{article}

\usepackage{PRIMEarxiv}

\usepackage[utf8]{inputenc} 
\usepackage[T1]{fontenc}    
\usepackage{hyperref}       
\usepackage{url}            
\usepackage{booktabs}       
\usepackage{amsfonts}       
\usepackage{nicefrac}       
\usepackage{microtype}      
\usepackage{lipsum}
\usepackage{fancyhdr}       
\usepackage{graphicx}       
\usepackage{float}

\usepackage{tikz}
\usetikzlibrary{calc,trees,positioning,arrows,chains,shapes.geometric,%
    decorations.pathreplacing,decorations.pathmorphing,shapes,%
    matrix,shapes.symbols}

\tikzset{
>=stealth',
  punktchain/.style={
    rectangle, 
    rounded corners, 
    draw=black, very thick,
    text width=15em, 
    minimum height=3em, 
    text centered, 
    on chain},
  line/.style={draw, thick, <-},
  element/.style={
    tape,
    top color=white,
    bottom color=blue!50!black!60!,
    minimum width=8em,
    draw=blue!40!black!90, very thick,
    text width=10em, 
    minimum height=3.5em, 
    text centered, 
    on chain},
  every join/.style={->, thick,shorten >=1pt},
  decoration={brace},
  tuborg/.style={decorate},
  tubnode/.style={midway, right=2pt},
}

\definecolor{ForestGreen}{RGB}{34,139,34}

\graphicspath{{media/}}     

\title{Bias and unfairness in machine learning models: a systematic literature review
}

\author{
Tiago Palma Pagano\textsuperscript{1}
Rafael Bessa Loureiro\textsuperscript{1},
Fernanda Vitória Nascimento Lisboa\textsuperscript{1} \\ 
\textbf{Gustavo Oliveira Ramos Cruz\textsuperscript{1},}
\textbf{Rodrigo Matos Peixoto\textsuperscript{1},}
\textbf{Guilherme Aragão de Sousa Guimarães\textsuperscript{1}}, \\
\textbf{Lucas Lisboa dos Santos\textsuperscript{1}}, \textbf{Maira Matos Araujo\textsuperscript{1}},
\textbf{Marco Cruz\textsuperscript{2}} \\
\textbf{Ewerton Lopes Silva de Oliveira\textsuperscript{2}}, 
\textbf{Ingrid Winkler\textsuperscript{1}},
\textbf{Erick Giovani Sperandio Nascimento\textsuperscript{1, 3*}} \\ \\
\textsuperscript{1}SENAI CIMATEC, Salvador - BA\\
\textsuperscript{2}HP Inc., Porto Alegre - RS\\
\textsuperscript{3}Surrey Institute for People-Centred AI, School of Computer Science and Electronic Engineering, \\ Faculty of Engineering and Physical Sciences, University of Surrey, Guildford, GU2 7XH, United Kingdom\\
\textit{\textsuperscript{*}Corresponding author: erick.sperandio@fieb.org.br, erick.sperandio@surrey.ac.uk } \\
}

\begin{document}
\maketitle

\begin{abstract}
One of the difficulties of artificial intelligence is to ensure that model decisions are fair and free of bias. In research, datasets, metrics, techniques, and tools are applied to detect and mitigate algorithmic unfairness and bias. This study aims to examine the latest existing knowledge about bias and unfairness in machine learning (ML) models with the RSL methodology and a bibliometric analysis. A Systematic Literature Review found 45 eligible articles published between 2017 and 2022 in the Scopus, IEEE Xplore, Web of Science, and Google Scholar knowledge bases. The results show numerous bias and unfairness detection and mitigation approaches for ML technologies, with clearly defined metrics in the literature, and varied metrics can be highlighted, we also address bias mitigation techniques, supporting tools, as well as more common datasets involving bias and unfairness identification and mitigation work, with binary and multi-class targets.
\end{abstract}


\section{Introduction}\label{sec1}

In industry, prediction-based decision algorithms are widely utilized by governments and organizations that are rapidly embracing them \cite{dwivedi2021artificial}. These techniques are already commonly used in lending, contracting, and online advertising, as well as in criminal pre-trial proceedings, immigration detention, and public health, among other areas \cite{mitchell2021algorithmic}. With the rise of these techniques, worry emerged about the biases embedded in the models and how fair they are, defining their performance for issues related to sensitive social aspects such as race, gender, class, etc \cite{verma2018fairness}.


Systems that have an influence on people's lives raise ethical concerns about making judgments in a fair and unbiased fashion. As a result, bias and unfairness challenges have been extensively investigated taking into account the limits imposed by corporate practices, regulations, social traditions, and ethical obligations. \cite{jones2020characterising}. Recognizing and reducing bias and unfairness are difficult tasks, since unfairness is defined differently in different cultures. As a consequence, user experience, cultural, social, historical, political, legal, and ethical considerations all have an impact on the unfairness criterion \cite{mitchell2019model}.


A definition for justice was given by \cite{di2022recommender}, according to the author injustice is defined as "systematic and unfair discrimination or prejudice of certain individuals or groups of individuals in favor of others." He further states that injustice is commonly caused by social or statistical biases, the former referring to the divergence between how the world should be and how it actually is, the latter is the discrepancy between how the world is and how it is encoded in the system.

In \cite{booth2021integrating} he differentiated the concepts of justice and bias, pointing out that usually works in the area use the two terms interchangeably. He defined that justice is a social concept of value judgment, being subjective, varying across cultures and societies  and in the context of organizations such as schools, hospitals, or corporations. Biases, on the other hand, are related to systematic errors that alter human behaviors or judgments about others because of their membership in a group determined by distinguishing characteristics such as gender or age.




As a result, new approaches from data science, artificial intelligence (AI), and machine learning (ML) are needed to take into account the aforementioned constraints on algorithms \cite{schumann2020we}.


The challenge worsens if key technological applications do not yet have ML models associated with the explainability of the decisions made, or those can only be evaluated by the team that created them, which leaves researchers unable to obtain these explanations and conduct experiments \cite{ammar2019cyber}. Given the millions of parameters analyzed by the machine, obtaining a transparent algorithm is quite challenging. Another option is to interpret it without knowing each step taken by the algorithm \cite{seymour2018detecting}.


Analyzing bias and unfairness collaborates with model explainability, so explainability is intrinsic to the study. According to \cite{gade2019explainable} explainability involves (i) defining model explainability, (ii) formulating explainability tasks to understand model behavior and developing solutions to those tasks, and finally, (iii) designing measures to evaluate model performance. Note that analyzing bias and unfairness directly addresses these topics, just as explainability promotes transparency.



Some solutions, such as AIF360, \cite{bellamy2018ai}, FairLearn \cite{bird2020fairlearn}, Tensorflow Responsible AI \cite{mitchell2019model} \cite{wexler2019if} \cite{tenney2020language} and Aequitas \cite{saleiro2018aequitas} are specific tools for dealing with bias and injustice.



However, the development approach to identify and mitigate bias and unfairness in ML models is left entirely to the developer, who often does not have adequate knowledge about the problem and must also consider aspects of fairness as a key element for the quality of the final model, confirming the need for a methodology to help deal with the problem \cite{ammar2019cyber}.



Another challenge is that most existing solutions for mitigating bias and unfairness are one-off applications for a specific problem or use case (UC). There are numerous approaches to identifying bias and unfairness, known as fairness metrics, and this wide range makes it difficult to select the right evaluation criteria for the issue you want to mitigate \cite{nielsen2020practical, kordzadeh2022algorithmic}.



Given the contextualization done so far, this study aims to examine the most recent existing knowledge on bias and unfairness in machine learning (ML) models with the RSL methodology and bibliometric analysis.

The present work consists of the study involving 45 papers, as noted in Section \ref{sec:Literature_Search}. There are other literature review papers, such as \cite{le2022survey, pessach2022review, surveybias2019, bacelar2021monitoring, balayn2021managing, chouldechova2018frontiers, Suresh2019Framework}. However, our search did not consider papers prior to 2017, aiming to present a more recent overview of bias and unfairness in machine learning models, since, as shown in Figure \ref{FIG:Reference_Spectroscopy}, there was a spike in publications in the year 2018 and previously there was a smaller number, so papers after this period should present more recent overviews consistent with the most effective and less embryonic solutions of the topic. Given the above, it is important to point out that \cite{le2022survey, bacelar2021monitoring, pessach2022review, surveybias2019, chouldechova2018frontiers} have conducted their research without specifying a recent period, allowing already outdated surveys.


The work of \cite{le2022survey} deals exclusively with datasets for bias and unfairness studies, without delving into mitigation aspects, while in \cite{balayn2021managing} there is more unpacking of mitigation aspects, but the focus is on data management, highlighting the unfairness topics for this area compared to the others presented. While in \cite{pessach2022review} the focus is mainly on classification problems, highlighting that other technics are opportunities that should emerge in the coming years, in \cite{chouldechova2018frontiers} is reinforced that work on algorithmic justice concentrates on single classification tasks, this finding were also identified by our work, however still \cite{chouldechova2018frontiers} does not do an in-depth on bias mitigation methods, as well as \cite{Suresh2019Framework} does not cite recent papers, having only one paper from 2019 and all others prior to this period.



Whereas \cite{bacelar2021monitoring} performs a more simplified analysis of fairness metrics and mitigation techniques, without addressing issues related to reference datasets for studies in the area. In the work of \cite{surveybias2019} they demonstrate why fairness is an important issue with examples of the possible harms in the real world, and they also examined the definitions of fairness and bias already proposed by researchers in different fields, such as general machine learning, deep learning and natural language processing, however the methodology for selecting the papers was not specified.


In \cite{kordzadeh2022algorithmic} also reviews the literature related to algorithmic bias and makes recommendations for future research, however it focuses only on theoretical aspects of justice. It asserts that the mechanisms by which technology-driven biases translate into decisions and behaviors have been largely ignored. He brings definitions on how a context classifies, which can be individual, task, technology, organizational, and environmental, and can influence the perceptual and behavioral manifestations of bias in the model. His work seeks a reflection on the behavior of people impacted by model decisions in order to use them as an element of influence in model decisions. In \cite{di2022recommender} emphasizes that addressing fairness in recommender systems provide different approaches, and, many gaps considering different users, research, the most important including: gender, age, ethnicity or personality.


As a differential, our work promotes a careful search methodology for the selection of the papers contained in Table \ref{tab:selected_works1}, in order to extract the concepts and techniques discussed in the period when the theme was most debated among the scientific community.



The paper primarily seeks, methods of bias and unfairness identification and mitigation for ML technologies, through fairness metrics, bias mitigation techniques, supporting tools, as well as more common datasets, involving work addressing bias and unfairness identification and mitigation, with binary and multiclass targets. The unfolding of each of these elements will be addressed in the following sections.



This paper is organized as follows: Section II describes the research method and the advantages of using RSL, Section III examines the results and addresses elements such as the Types of Bias, the Identified Datasets with the main problems for identifying and reducing bias and unfairness, the Justice Metrics for measuring the models bias and unfairness in different ways, and from the identification of bias it is possible to approach the Techniques and models for bias and unfairness Mitigation, either by manipulating the data (pre-processing), the model itself (in-processing) or the prediction (post-processing), some techniques mainly of in-processing promote the identification of the sensitive attribute, important to train models independent of them. Section IV presents our final considerations and suggestions for further research.

\section{Method}
\label{sec:Literature_Search}
A Systematic Literature Review (SLR) aims to consolidate research by bringing together elements for understanding it \cite{jones2020characterising}. Literature reviews are a widely used methodology to gather existing findings into a research field \cite{kraus2020art}. 

This systematic review followed the Preferred Reporting Items for Systematic reviews and Meta-Analyses (PRISMA) guidelines \cite{page2021prisma} and was conducted using the method described in  \cite{pagano2022machine} and in \cite{booth2016systematic}, which encompasses five steps: Planning, Scoping, Searching, Assessing, Synthesizing.



During the Planning step, the knowledge bases that will be explored are defined \cite{booth2016systematic}. The search for document patents was undertaken in the following knowledge bases:


\begin{itemize}
	\item IEEE Xplore (www.ieeexplore.ieee.org/)
	\item Scopus (www.periodicos.capes.gov.br)
	\item Web Of Science (www.periodicos.capes.gov.br/)
	\item Google Scholar (www.scholar.google.com)
\end{itemize}

These bases were chosen because they are reliable and multi-disciplinary knowledge databases of international scope, with comprehensive coverage of citation indexing, allowing the best data from scientific publications.



The Scope Definition step ensures that questions relevant to the research are considered before the actual Literature Review is carried out \cite{booth2016systematic}. A brainstorming session was held with an interdisciplinary group composed of eleven experts on machine learning models, which selected two pertinent research questions to this systematic review address, namely: 


Q1: What is the state of the art on the identification and mitigation of bias and unfairness in ML models?


Q2: What are the challenges and opportunities for identifying and mitigating bias and unfairness in ML models?



The Literature Search involves exploring the databases specified in the Planning step in a way that aims to solve the questions defined in the scope \cite{booth2016systematic}.

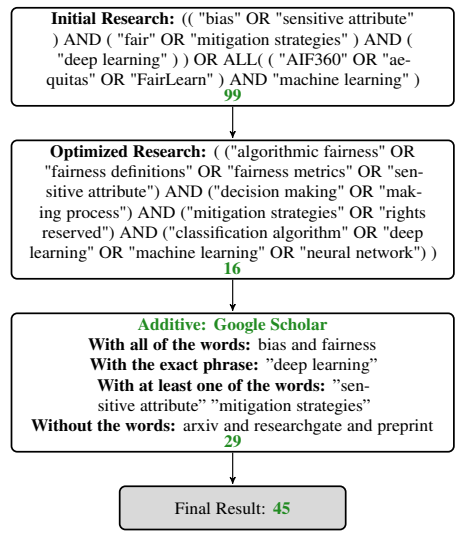
\begin{figure}
\centering
\resizebox{6cm}{!}
{
\begin{tikzpicture}
  [node distance=.8cm,  start chain=going below,minimum size=5cm]
     \node[punktchain, join, text width=30em] (intro) {\large{\textbf{Initial Research:} (( "bias"  OR  "sensitive attribute" )  AND  ( "fair"  OR  "mitigation strategies" )  AND  ( "deep learning" ) ) OR ALL( ( "AIF360"  OR  "aequitas" OR  "FairLearn" )  AND  "machine learning" )\\\textcolor{ForestGreen}{\textbf{99}}}             };
     
     \node[punktchain, join, text width=30em] (probf)      {\large{\textbf{Optimized Research:} ( ("algorithmic fairness" OR "fairness definitions" OR "fairness metrics" OR "sensitive attribute") AND  ("decision making" OR "making process") AND  ("mitigation strategies" OR "rights reserved") AND     ("classification algorithm" OR "deep learning" OR "machine learning" OR "neural network") )\\\textcolor{ForestGreen}{\textbf{16}}}};
     
     \node[punktchain, join, text width=30em] (investeringer)      {\large{\textcolor{ForestGreen}{\textbf{Additive: Google Scholar}}\\           \textbf{With all of the words:} bias and fairness\\                            \textbf{With the exact phrase:} ''deep learning''\\                         \textbf{With at least one of the words:} ''sensitive attribute'' ''mitigation strategies''\\
     \textbf{Without the words:} arxiv and researchgate and preprint\\ \textcolor{ForestGreen}{\textbf{29}}}        };
     
     \node[punktchain, join, fill=gray!30] (perfekt) {\large{Final Result:  \textcolor{ForestGreen}{\textbf{45}}}};
  \end{tikzpicture}
}
\caption{Papers Selection Process with the number of papers}
\label{FIG:Studies_selection_process}
\end{figure}

Initially, the keywords were used to search the knowledge bases noted in Figure \ref{FIG:Studies_selection_process}. In addition to studies on bias or sensitive attributes using fairness or mitigation strategies for machine learning, it should include studies using the ''AIF360'', ''Aequitas'' or ''FairLearn'' tools for ML. This inclusion in the initial search aims to relate tools for identifying and mitigating bias and unfairness to the optimized search criteria, including the most important tools in the literature. These criteria defined the 'initial search', with 99 publications selected. Only review papers, research papers and conferences were considered.


These papers were used to optimize the search criteria using the litsearchr \cite{grames2019automated} library, which assembles a word co-occurrence network to identify the most relevant words. The optimized search yielded 16 selected papers, which can also be seen in Figure \ref{FIG:Studies_selection_process}.


In addition, a Google Scholar search was performed, and 29 publications were selected based on their Title and Abstract fields. The search was based on the string used in the databases, applying the same keywords in the advanced search criteria, as can be seen in Figure \ref{FIG:Studies_selection_process}. The search in Google Scholar aims to select papers that might not have been indexed in the knowledge bases.

The Assessing the Evidence Base step selects the most relevant articles based on bibliometric analysis and reading the article abstracts.

\begin{figure*}[h]
	\centering
		\includegraphics[scale=0.28]{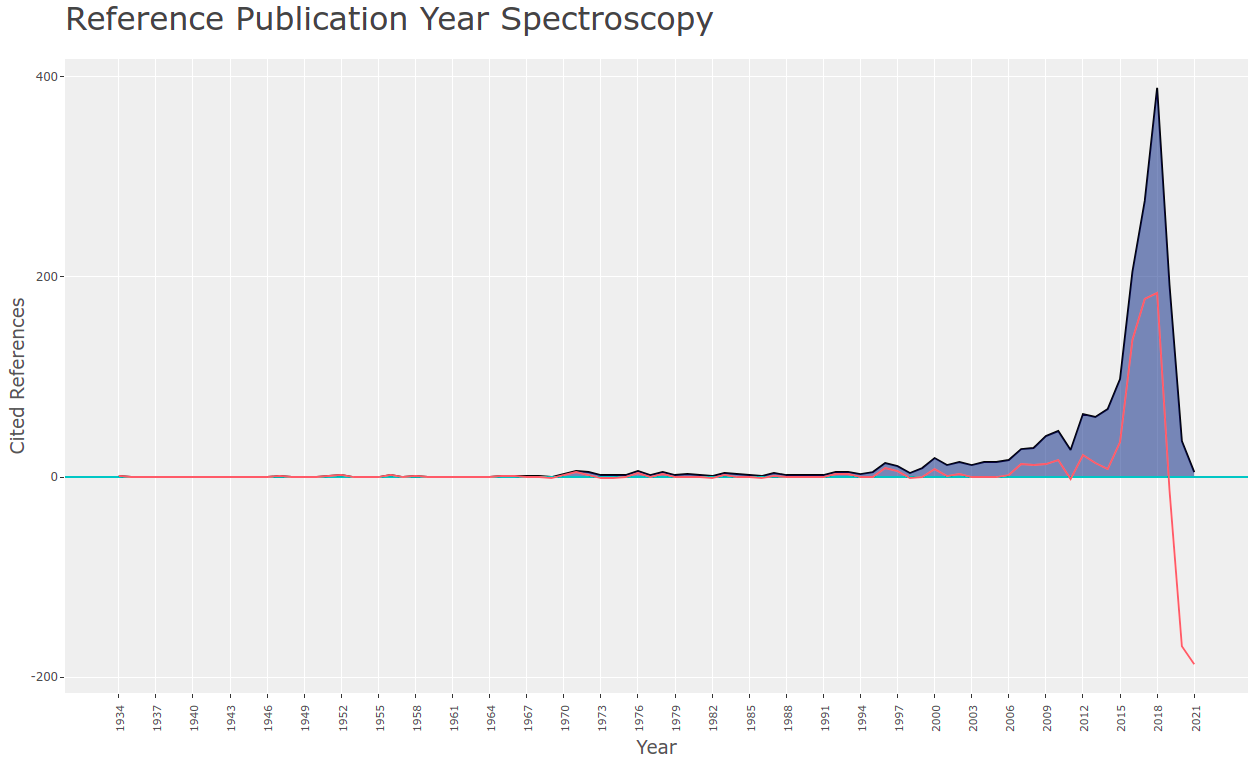}
	\caption{Year of the references cited in the papers}
	\label{FIG:Reference_Spectroscopy}
\end{figure*}

Initially, searches in the four knowledge bases retrieved 99 articles, with the fields Title, Abstract, and Keywords serving as search criteria. We included only Review Articles, Research Articles, and Conference Proceedings published between 2017 and 2022, as shown by the bibliometric analysis in Figure \ref{FIG:Reference_Spectroscopy}. The red line represents the average difference in the number of articles over the previous five years, with a decrease in the final year due to the time span covered by the search.

\begin{figure*}[h]
	\centering
		\includegraphics[scale=0.30]{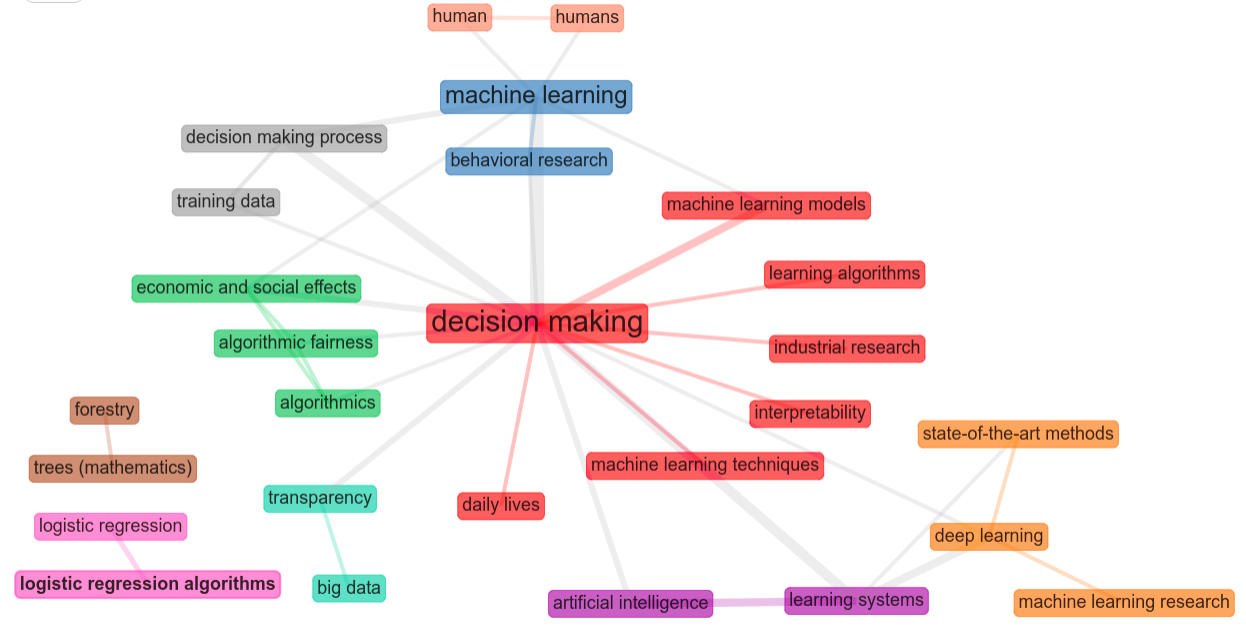}
	\caption{Keyword co-occurrence network}
	\label{FIG:Co-occurrence_Network}
\end{figure*}

A graph of the relationship between the keywords obtained in the search was generated using the biblioshiny tool \cite{patil2020global} from the bibliometrix package \cite{aria2017bibliometrix} in the R language. Figure \ref{FIG:Co-occurrence_Network} illustrates some clusters that exemplify themes addressed in the RSL papers. The red cluster relates to "machine learning" and decision-making in models, the green cluster considers "fairness" and its economic and social impacts. It is also worth highlighting aspects related to transparency, interpretability, and the relationship of these keywords with the state of the art.


As a result of the final search, 45 articles were selected for discussion as shown in Figure \ref{FIG:Studies_selection_process}.


The Synthesis and Analysis step consists in reading and evaluating the selected articles to identify patterns, differences, and gaps that might be studied further in future research on bias and unfairness in machine learning models.


\section{Results}

This section presents and analyses the 45 selected studies, which are included in Table \ref{tab:selected_works1}, according to the research questions Q1 and Q2 set in the Scope Definition step.
The results are organized in five sections: Types of Bias, Identified Datasets, Mitigation techniques and models,
Technique for identification of the sensitive attribute and Fairness metrics. Those sections represent fundamental aspects of the discussion of bias and fairness.


The studies examined revealed issues that support the concern about bias and fairness in ML models. \cite{ammar2019cyber} addresses issues such as the lack of transparency of ML models, organizations such as Facebook and Telegram's lack of commitment to revealing the measures being taken in this effort, and even the constraints of resources, whether human or computer. 



In the same manner, \cite{di2022recommender} highlights the importance of responsible AI, although there is still no clear and globally accepted definition of responsibility for AI systems. This should include fairness, security and privacy, explainability, safety and reproducibility. Specifically on fairness it highlights that the regulation should emphasize obligations to "... minimize the risk of erroneous or biased decisions in critical areas such as education and training, employment, important services, law enforcement, and the judiciary." 

Initially \cite{ammar2019cyber}, criticizes the complexity of comprehending ML models, which can only be examined by the team that developed them, and which frequently does not understand all of the model's features or why it made certain judgments. Furthermore, the more complex the model, the more difficult it is to analyze its decision-making process. 


The study  \cite{kozodoi2022fairness} aims to provide an overview and a systemic view regarding recent criteria and processes in machine learning development, and to conduct empirical tests on the use of these for credit scores. The authors selected which fairness criteria best fit for these scores and cataloged state-of-the-art fairness processors, using them to identify when loan approval processes are met. Using seven datasets of credit scores, they performed empirical comparisons for different fairness processors.


The \cite{stoyanovich2020responsible} study found security and transparency issues with automated decision systems (ADS), warning and urging data engineers to develop a more fair and inclusive procedure. For the authors, ADS must be accountable in the following areas: development, \textit{design}, application, and usage, as well as rigorous regulation and monitoring, so that they do not perpetuate inequality.




The advantages and disadvantages of transparency in machine learning models, defining bias, fairness and arguing that a transparent algorithm is extremely difficult to obtain given the millions of parameters analyzed by a machine \cite{seymour2018detecting}. An alternative is a transparent output that can be analyzed and understood without having to understand every step made by the algorithm. To define transparency in \cite{seymour2018detecting} two categories have been defined: process transparency and result transparency.

The term "process transparency" refers to an understanding of the algorithm's underlying characteristics, such as the attributes it weighs in its decisions. The term "result transparency" refers to the capacity to understand decisions and patterns in classification process answers. In addition, the model must meet two requirements: global and local explanation. The Local explanation includes a detailed examination of which characteristics were most important in reaching a particular decision, whereas the Global explanation evaluates all decisions based on certain metrics. The author suggests a mental model of the main system for this evaluation, and if it can predict what the classification of the main model is, it is on the correct course to transparency. Finally, it is stated that a premise of \textit{white-box} and \textit{black-box} models might bring out implicit and explicit features of the models and facilitate auditors' job \cite{seymour2018detecting}.


ML models, whether classification or regression, can be of type \textit{White-box} or \textit{Black-box}, depending on their availability and constraints:


\begin{itemize}
\item \textbf{\textit{White-box}}: these are machine learning models that deliver easy to understand outcomes for application domain specialists. Typically, these models provide a good tradeoff between accuracy and explainability \cite{loyola2019black} and hence have less constraint and difficulties for structural adjustments. The structure and functioning of this model category are simple to grasp.

\item \textbf{\textit{Black-box}}: ML models that, from a mathematical perspective, are extremely difficult to explain and comprehend by specialists in practical areas \cite{loyola2019black}. Changes to the structure of models in this category are restricted, and it is difficult to grasp their structure and functioning.

\end{itemize}

\begin{table*}
    \centering
\hspace*{-0.8cm}
\begin{tabular}{p{.5pc}p{38pc}}
\toprule
& Study \\
\midrule

1 & \footnotesize{One-Network Adversarial Fairness \cite{adel2019one}}        \\ \hline

2 & \footnotesize{Cyber gremlin: Social networking, machine learning and the global war on al-qaida-and is-inspired terrorism \cite{ammar2019cyber}}         \\ \hline

3 & \footnotesize{Fairness research on deep learning \cite{jinyin2021fairness}}         \\ \hline

4 & \footnotesize{Mdfa: Multi-differential fairness auditor for black box classifiers \cite{gitiaux2019mdfa}}          \\ \hline

5 & \footnotesize{Algorithmic fairness: Choices, assumptions, and definitions \cite{mitchell2021algorithmic}}          \\ \hline

6 & \footnotesize{Dynamic fairness – breaking vicious cycles in automatic decision making  \cite{paassen2019dynamic}}         \\ \hline

7 & \footnotesize{Recycling privileged learning and distribution matching for fairness \cite{quadrianto2017recycling}}         \\ \hline

8 & \footnotesize{We need fairness and explainability in algorithmic hiring blue sky ideas track \cite{schumann2020we}}          \\ \hline

9 & \footnotesize{Detecting bias: Does an algorithm have to be transparent in order to be fair? \cite{seymour2018detecting}}         \\ \hline

10 & \footnotesize{Fairness-Aware Instrumentation of Preprocessing Pipelines for Machine Learning \cite{yang2020fairness}}        \\ \hline

11 & \footnotesize{Improving machine learning fairness with sampling and adversarial learning \cite{amend2021improving}}    \\ \hline

12 & \footnotesize{Responsible data management \cite{stoyanovich2020responsible}}         \\ \hline

13 & \footnotesize{Using Machine Learning in Admissions: Reducing Human and Algorithmic Bias in the Selection Process \cite{martinez2021using}}   \\ \hline

14 & \footnotesize{Analysis bias in sensitive personal information used to train financial models \cite{bryant2019analyzing}}    \\ \hline

15 & \footnotesize{VITAL-ECG: A de-bias algorithm embedded in a gender-immune device \cite{paviglianiti2020vital}}        \\ \hline

16 & \footnotesize{Dataset bias: A case study for visual question answering \cite{das2019dataset}}       \\ \hline

17 & \footnotesize{A causal bayesian networks viewpoint on fairness \cite{chiappa2018causal}}        \\ \hline

18 & \footnotesize{Constraining deep representations with a noise module for fair classification \cite{cerrato2020constraining}}         \\ \hline

19 & \footnotesize{Algorithm Bias Detection and Mitigation in Lenovo Face Recognition Engine \cite{shi2020algorithm}}       \\ \hline

20 & \footnotesize{Artificial Intelligence (AI): Multidisciplinary perspectives on emerging challenges, opportunities, and agenda for research, practice and policy \cite{dwivedi2019artificial}}      \\ \hline

21 & \footnotesize{Cost-sensitive hierarchical classification via multi-scale information entropy for data with an imbalanced distribution \cite{zheng2021cost}}        \\ \hline

22 & \footnotesize{Fair adversarial gradient tree boosting \cite{grari2019fair}}      \\ \hline

23 & \footnotesize{Harnessing artificial intelligence (AI) to increase wellbeing for all: The case for a new technology diplomacy \cite{feijoo2020harnessing}}       \\ \hline

24 & \footnotesize{Privacy and Ethical Challenges in Big Data \cite{gambs2018privacy}}   \\ \hline

25 & \footnotesize{Singular race models: Addressing bias and accuracy in predicting prisoner recidivism \cite{jain2019singular}}     \\ \hline

26 & \footnotesize{Fairness in Credit Scoring: Assessment, Implementation and Profit Implications \cite{kozodoi2022fairness}}   \\ \hline

27 & \footnotesize{When Politicization Stops Algorithms in Criminal Justice \cite{konig2021politicization}}       \\ \hline

28 & \footnotesize{A survey on bias and fairness in machine learning,  \cite{Mehrabi2021}}       \\ \hline

29 & \footnotesize{Evolution and impact of bias in human and machine learning algorithm interaction  \cite{sun2020evolution}}  \\ \hline

30 & \footnotesize{Benchmarking Bias Mitigation Algorithms in Representation Learning through Fairness Metrics \cite{reddy2021benchmarking}} \\ \hline

31 & \footnotesize{Fairness for image generation with uncertain sensitive attributes \cite{jalal2021fairness}}    \\ \hline

32 & \footnotesize{Mitigating Demographic Bias in Facial Datasets with Style-Based Multi-attribute Transfer \cite{georgopoulos2021mitigating}}        \\ \hline

33 & \footnotesize{Constructing a Fair Classifier with the Generated Fair Data \cite{jang2021constructing}} \\ \hline

34 & \footnotesize{Enforcing fairness in logistic regression algorithm  \cite{radovanovic2020enforcing}}     \\ \hline

35 & \footnotesize{Fairness metrics and bias mitigation strategies for rating predictions \cite{ashokan2021fairness}}    \\ \hline

36 & \footnotesize{Fairness via Representation Neutralization \cite{du2021fairness}}  \\ \hline

37 & \footnotesize{Improving fairness of artificial intelligence algorithms in Privileged-Group Selection Bias data settings \cite{pessach2021improving}}     \\ \hline

38 & \footnotesize{Fairness in Deep Learning: A Computational Perspective \cite{Du2021}}   \\ \hline

39 & \footnotesize{Risk Identification Questionnaire for Detecting Unintended Bias in the Machine Learning Development Lifecycle  \cite{lee2021risk}}  \\ \hline

40 & \footnotesize{Fair Outlier Detection Based on Adversarial Representation Learning  \cite{li2022fair}} \\ \hline

41 & \footnotesize{Recommender Systems under European AI Regulations  \cite{di2022recommender}} \\ \hline

42 & \footnotesize{Psychological Measurement in the Information Age: Machine-Learned Computational Models  \cite{d2022psychological}} \\ \hline

43 & \footnotesize{Monitoring Fairness in HOLDA  \cite{fontanamonitoring}} \\ \hline

44 & \footnotesize{Algorithmic bias: review, synthesis, and future research directions  \cite{kordzadeh2022algorithmic}} \\ \hline

45 & \footnotesize{Integrating Psychometrics and Computing Perspectives on Bias and Fairness in Affective Computing: A case study of automated video interviews \cite{booth2021integrating}} \\
\bottomrule
\end{tabular}
\hspace*{-1cm}
\caption{Papers returned by the search string}
\label{tab:selected_works1}
\end{table*}

The work \cite{mitchell2021algorithmic} corroborates the argumentation of \cite{ammar2019cyber}, emphasizing that when dealing with people, even the finest algorithm will be biased if sensitive attributes are not taken into consideration. One of the first issues raised is that the prejudice and justice literature is often confined to addressing the situation of a group or individual experiencing injustice in the present time. In this case, one must broaden the search and analyze how the individual's effect impacts his or her community and vice versa. Dataset and people behavior is fluid and can diverge dramatically over a few years, but the algorithm may retain a bias in its training and be unable to adapt to this shift. A group that is mistreated in the actual world would almost certainly be wronged by the algorithm, and that this type of bias just reflects reality rather than being a biased dataset.

In the work of \cite{d2022psychological} a problem is raised in the example of a model that evaluates the personality of patients using automated video interviews. If men have higher scores than women, this could be considered bias. On the other hand, if the annotations for agreement indicate higher scores for men than for women, the model reproduces this pattern and cannot be considered biased. There is an ambiguity in that the model would be fair because its measures reflect observable reality, while at the same time unfair because it gives unequal results to the group. This confusion occurs because of the lack of knowledge about the identification of the model's group bias, which uses right and wrong predictions criteria on the target provided by the dataset. Therefore, the model's right and wrong rates should be the same for different groups. There are numerous relationships that use these rates, as will be seen below.

With a similar opinion, \cite{konig2021politicization} opposes the use of models for decision-making, defining the use of tools for risk assessment in models for pre-judgment as a justification. The authors argue that the implementation of these tools can introduce new uncertainties, disruptions, and risks into the judgment process. By conducting empirical experiments with unfair models, they conclude that the process of implementing these tools should be stopped.


Furthermore, \cite{paassen2019dynamic} states that while there are various fair models for classification tasks, these are restricted to the present time, and because they  embed the human bias, there is a propensity to repeat and escalate the segregation of particular groups through a vicious cycle. Whereas a classifier that gives a group a higher number of good ratings will give it an advantage in the future, and vice versa for negative ratings.


Meanwhile, \cite{ammar2019cyber} also claims that algorithms frequently disregard uncommon information, framing the act as censorship, such as Islamism and terrorist content. Because of this issue, decision-making algorithms tend to be biased toward more common occurrences in their case-specific databases. 


Finally, \cite{dwivedi2019artificial} brings together the perspectives of various experts, emphasizing opportunities from the usage of AI, evaluating its impact, challenges, and the potential research agenda represented by AI's rapid growth in various fields of industry and society in general.  Tastes, anxieties, and cultural proximity seem to induce bias in consumer behavior, which will impact demand for AI goods and services, which is, according to the study, an issue that is yet under research.


Inferring patterns from large datasets in an unbiased environment and developing theories to explain those patterns can eliminate the need for hypothesis testing, eradicating the bias in the analysis data and, consequently, in the decisions \cite{dwivedi2021artificial}.

In the \cite{dwivedi2019artificial} paper, general issues around ML are addressed, where governments are increasingly experimenting with them to increase efficiency in large-scale personalization of services based on citizen profiles, such as predicting viral outbreaks and crime hotspots, and AI systems used for food safety inspections. Bias in this context implicates governance issues, which pose dangers to society because algorithms can develop biases that reinforce historical discrimination, undesirable practices, or result in unexpected effects due to hidden complexities. Other related themes include ethics, transparency and audits, accountability and legal issues, fairness and equity, protection from misuse, and the digital divide and data deficit. These aspects are reinforced by \cite{feijoo2020harnessing} when it states that the discussion should expand to include technology diplomacy as a facilitator of global policy alignment and governance, for developing solutions to avian flu, for example. It also discusses the importance of implementing fundamental ethical concepts in AI, such as beneficence, non-maleficence, decision-making, fairness, explainability, reliable AI, suggested human oversight, alternative decision plans, privacy, traceability, non-discrimination, and accountability. 

The work \cite{gambs2018privacy} addresses the issue of data privacy, as well as other ethical challenges related to Big Data research, such as transparency, interpretability, and fairness of algorithms based on this data. It is critical to explore methods to assess and quantify the bias of algorithms that learn from Big Data, particularly in terms of potential dangers of discrimination against population subgroups, and to suggest strategies to rectify unwarranted bias. 


It also deals with the difference between individual justice and group justice, where the former states that individuals who are similar except for the sensitive attribute should be treated similarly and given similar decisions. This relates to the legal concept of unequal treatment when the decision-making process is based on sensitive attributes. However, individual justice is only relevant when the decision-making process causes discrimination, and cannot be used when the goal is to address biases in the data. Group justice, on the other hand, depends on the statistics of the outcomes of the subgroups indexed in the data and can be quantified in various ways, such as demographic parity and equalized odds, and thus can have bias addressed in the data \cite{gambs2018privacy}. In \cite{booth2021integrating} states that group fairness, considers that groups contain useful information to adjust predictions, making them more accurate, highlighted that the metrics statistical parity, group fairness, and adverse impact are all concerned with equality of acceptance rates across groups.

\subsection{Types of Bias}


The types of biases are the pre-existing ones when they exist independently of an algorithm itself and have their origins in society. The technical bias, on the other hand, is occurring because of the systems developed, and it can be treated, measured, and its cause understood. He also defined the type of bias called emergent, which occurs when a system is designed for different users or when social concepts change \cite{stoyanovich2020responsible}.


For \cite{Mehrabi2021} bias can be classified into data bias, algorithm bias, and user interaction. The first considers that bias is present in the data, such as unbalanced data for example. The second one addresses the bias caused exclusively by the algorithm, caused by optimization functions, regularization, among other causes. The third type of bias is caused by the interaction with the user, since the interface allows the user to impose his/her behavior for a self-selected biased interaction.


The work \cite{sun2020evolution} brings the concept of iterated algorithmic bias, features present in recommendation systems, with the types: filtering bias, the active learning bias, and the random-based bias. The first occurs when the goal is to provide relevant information or preferences. The second occurs when it aims to predict with the user's preferences. And the last one is based on an unbiased approach and used as a baseline for no user preference.

The study \cite{stoyanovich2020responsible} defines three types of bias: pre-existing, technological, and emergent. The pre-existing bias category refers to data that reflects inequalities absorbed by the algorithm, hence spreading them. The technical category relates to bias worsening pre-existing prejudice caused by one of the algorithm's internal decision processes, and this may be addressed rather simply in comparison to the others. Finally, the emergent category refers to bias that occurs as a result of the usage of one or more users. For example, if a manager assigns higher performance to male employees, the algorithm is likely to begin favoring them and/or incorrectly rating women in the same division of the organization.


The paper \cite{Mehrabi2021} goes further in this concept, listing the 23 most common sources of bias, and these are divided into three categories organized in order to consider the feedback loop, they are: data, algorithm, and user interaction. So we have some examples of biases:


\begin{itemize}
\item Historical and social: coming from the data;
\item Emerging and popularity: coming from the algorithm;
\item Behavioral and presentation bias: caused by interaction with the user.
\end{itemize}




In the article \cite{sun2020evolution} the authors proposed a framework to analyze bias and concluded that filtering bias, prominent in personalized user interfaces, can limit the discoverability of relevant information to be presented. In addition, they address the importance and damage caused by feedback loops and how algorithm performance and human behavior influence each other by denying certain information to a user, impacting long-term performance.


The work \cite{lee2021risk} proposed a methodology to identify the risks of potential unintended and harmful biases in ML. They therefore developed a practical risk assessment questionnaire to identify the sources of bias that cause unfairness and applied it to cases such as criminal risk prediction, health care provisions, and mortgage lending. The questionnaire was validated with industry professionals, and 86\% agreed it was useful for proactively diagnosing unexpected issues that may arise in the ML model. Note that this work allows you to identify causes that may bias the models in a theoretical way.

\subsection{Identified Datasets}

A survey of the datasets used in the papers was conducted, listed in Table \ref{tab:datasets_fairness}. These datasets mostly are known to include demographic annotations, allowing to assess unfairness and bias in their data, and can be used to test and validate techniques aimed at resolving these issues. Other datasets do not have demographic data, as it aims to identify bias and unfairness in image generation, reconstruction, enhancement, and super-resolution, not necessarily associated with demographic sensitive issues \cite{jalal2021fairness}.

\begin{table*}[h]
\setlength{\tabcolsep}{18pt}
\renewcommand{\arraystretch}{0.4}
 \centering
\begin{tabular}{p{8em}p{35em}}
\toprule
Metrics  & References \\ 
\midrule
COMPAS & \cite{jinyin2021fairness, yang2020fairness, jang2021constructing, adel2019one, quadrianto2017recycling, cerrato2020constraining, radovanovic2020enforcing, gitiaux2019mdfa, li2022fair}  \\ \hline
Communities & \cite{gitiaux2019mdfa} \\ \hline
FDOC  & \cite{jain2019singular} \\ \hline
FDLE  & \cite{jain2019singular} \\ \hline

Student  & \cite{li2022fair} \\ \hline

Bank & \cite{jinyin2021fairness, bryant2019analyzing, cerrato2020constraining} \\ \hline
German & \cite{jinyin2021fairness, jang2021constructing, cerrato2020constraining, gitiaux2019mdfa} \\ \hline
Credit  & \cite{li2022fair} \\ \hline
Adult & \cite{jinyin2021fairness, yang2020fairness, jang2021constructing, bryant2019analyzing,reddy2021benchmarking, adel2019one, quadrianto2017recycling, amend2021improving, cerrato2020constraining, radovanovic2020enforcing, du2021fairness, li2022fair,gitiaux2019mdfa}  \\ \hline
Boston & \cite{jinyin2021fairness} \\ \hline
MEPS & \cite{jinyin2021fairness, jang2021constructing, du2021fairness} \\ \hline
Heart & \cite{jinyin2021fairness} \\ \hline
MIMIC II  & \cite{paviglianiti2020vital} \\ \hline
Weight  & \cite{li2022fair} \\ \hline
Drug  & \cite{li2022fair} \\ \hline
AFHQ Cats and Dogs & \cite{jalal2021fairness} \\ \hline
LFW  & \cite{georgopoulos2021mitigating} \\ \hline
CelebA & \cite{georgopoulos2021mitigating, du2021fairness} \\ \hline
MOPRH  & \cite{georgopoulos2021mitigating} \\ \hline
MovieLens 1M  & \cite{ashokan2021fairness} \\ \hline
CI-MNIST & \cite{reddy2021benchmarking} \\ \hline
VQA  & \cite{das2019dataset} \\ \hline
VizWiz  & \cite{das2019dataset} \\ \hline
CLEVR  & \cite{das2019dataset} \\ \hline
Sintético  & \cite{gitiaux2019mdfa} \\ 

\bottomrule
\end{tabular}
\caption{Table with the datasets present in each paper.}
\label{tab:datasets_fairness}
\end{table*}

Some datasets address crime-related issues such as Propublica Correctional Offender Management Profiling for Alternative Sanctions (COMPAS), Communities and Crime (Communities), and Florida Department of Corrections (FDOC).

The COMPAS \cite{larson_mattu_kirchner_angwin_2016} dataset describes a binary classification task, which shows whether an inmate will reoffend within two years, has sensitive attributes such as race, age and gender. This is one of the most widely used datasets for bias and fairness experiments, with a controversial and relevant topic.

Similar in purpose to COMPAS, the Communities dataset \cite{Dua:2019} compares various socioeconomic situations of US citizens in the 1990s with the crime rate, identifying the per capita rate of violent crime in each community.

The FDOC \cite{jain2019singular} dataset, on the other hand, contains sentences with the types of charges, which can be violent charges (murder, manslaughter, sex crimes, and other violent crimes); robbery; burglary; other property charges (including theft, fraud, and damage); drug-related charges; and other charges (including weapons and other public order offenses). The dataset uses Florida Department of Law Enforcement (FDLE) criminal history records for recidivism information within 3 years. They have the attributes such as the major crime category, the offender's age of admission and release, time served in prison, number of crimes committed prior to arrest, race, marital status, employment status, gender, education level, and if recidivist whether or not they were supervised after release.

Addressing issues concerning the selection process and approval of individuals, the Student \cite{cortez2008using} dataset has the data collected during 2005 and 2006 in two public schools in Portugal. The dataset was built from two sources: school reports, based on sheets of paper including some tributes with the three grades of the period and number of school absences; and questionnaires, used to complement the previous information. It also includes demographic data with mother's education, family income, social/emotional situation, alcohol consumption, variables that can affect student performance.

Another theme found in the selected datasets involves financial issues of bank credit such as Bank marketing (Bank), German credit (German) and Credit. Wage forecasting with the Adult dataset and product pricing with the Boston housing price (Boston) dataset.

The Bank dataset is related to the marketing campaigns of a Portuguese bank between the years 2008 to 2013. The goal of the classification is to predict whether or not a customer will make a deposit subscription \cite{bryant2019analyzing}.

With similar purpose, the German \cite{Dua:2019} dataset has 1000 items and 20 categorical attributes. Each entry in this dataset represents an individual who receives credit from a bank. According to the set of attributes, each individual is evaluated on his or her credit risk. 

The Credit \cite{yeh2009comparisons} dataset, on the other hand, contains payment data from a Taiwanese bank (a cash and credit card issuer) for the purpose of identifying the bank's credit card holders who would potentially receive a loan. It has demographic annotations such as education level, age, and gender.

One of the most widely used datasets, Adult \cite{Dua:2019} has 32,561 full cases representing adults from the 1994 US census. The task is to predict whether an adult's salary is above or below \$50,000 based on 14 characteristics. The sensitive attribute 'gender' is embedded in the samples.

For real estate pricing, the Boston dataset has data extracted from the Boston Standard Metropolitan Statistical Area (SMSA) in 1970 and each of the 506 samples represents data obtained on 14 characteristics for households. The classification of this model aims to predict the property value of the region using attributes such as crime rate, proportion of residential land, average number of rooms per household, among others \cite{jinyin2021fairness}.

Another characteristic of the datasets found highlights applications in the health area, either to predict patients' financial expenses, as in the dataset Medical Expenditure Panel Survey (MEPS), or to identify possible health risks for patients as in the datasets: MEPS, Heart Disease (Heart), Multiparameter Intelligent Monitoring in Intensive Care (MIMIC II), Weight, and Drugs.

The MEPS \cite{creedon2022effects} dataset contains data on families and individuals in the United States, with their medical providers and employers, with information on the cost and use of health care or insurance.

To identify and prevent diseases the Heart \cite{jinyin2021fairness} dataset contains 76 attributes, but all published experiments refer to the use of a subset of 14 of them. The target attribute refers to the presence of heart disease in the patient and can be 0 (no presence) to 4. Experiments aim to classify the presence or absence of heart disease.

In the same vein as Heart, the MIMIC II \cite{paviglianiti2020vital} dataset contains vital signs captured from patient monitors and clinical data from tens of thousands of Intensive Care Unit (ICU) patients. It has demographic data such as patient gender and age, hospital admissions and discharge dates, room tracking, dates of death (in or out of hospital), ICD-9 codes, unique code for healthcare professional and patient type, as well as medications, lab tests, fluid administration, notes and reports.

The Weight \cite{de2019obesity} dataset contains data for estimating obesity levels in individuals from the countries of Mexico, Peru, and Colombia, based on their eating habits and physical condition. It has 17 attributes and 2,111 samples, labeled with the level of obesity which can be Low Weight, Normal Weight, Overweight Level I, Overweight Level II, Obesity Type I , Obesity Type II and Obesity Type III. The sensitive attributes are gender, age, weight, height, smoking, among others.

To predict narcotic use, the Drug \cite{fehrman2017five} dataset was collected from online survey including personality traits (NEO-FFI-R), impulsivity (BIS-11), sensation seeking (ImpSS), and demographic information. The \textit{dataset} contains information on the use of 18 central nervous system psychoactive drugs such as amphetamines, cannabis, cocaine, ecstasy, legal drugs, LSD, and magic mushrooms, among others. It has demographic attributes such as gender, education level, and age group.

In the area of image enhancement and face recognition, bias may not be associated with demographic features, the datasets that have demographic information were identified, among them: Labeled Faces in the Wild (LFW), Large-scale CelebFaces Attributes (CelebA), MORPH Longitudinal Database (MORPH), MovieLens 1M and Visual Question Answering (VQA). The dataset Animal FacesHQ (AFHQ) deals with the identification of animals, and the bias is associated with implicit features of the images, as well as the dataset Correlated and Imbalanced MNIST (CI-MNIST). Synthetic datasets were also found as an alternative.

The LFW \cite{georgopoulos2021mitigating} dataset contains 13,233 images of faces of 5749 distinct people and 1680 individuals are in two or more images. LFW is applied to face recognition problems and the images were annotated for demographic information such as gender, ethnicity, skin color, age group, hair color, eyeglass wearing, among other sensitive attributes.

The CelebA \cite{liu2015faceattributes} dataset contains 202,599 face images with 10,177 individuals and 40 annotated attributes per image such as gender, Asian features, skin color, age group, head color and eye color, among other sensitive attributes, just as LFW is also used for face recognition problems.

The MORPH dataset contains over 400,000 images of almost 70,000 individuals. The images are 8-bit color and sizes can vary. MORPH has annotations for age, sex, race, height, weight, and eye coordinates.

The MovieLens 1M \cite{ashokan2021fairness} dataset contains a set of movie ratings from the MovieLens website, a movie recommendation service of 1 million reviews from 6,000 users for 4,000 movies, with demographics such as gender, age, occupation, and zip code, plus data from the movies and the ratings.

The VQA \cite{das2019dataset} dataset contains natural language questions about images. It has 250,000 images, 760,000 questions and about 10 million answers. The questions have a sensitive criterion from the point of view of the questioner, and can be a simple question or a very difficult one, creating a bias. The images can also be very complex, making it difficult to identify the question element. The VizWiz dataset has the same proposal as the VQA for object recognition and assistive technologies, collected from users with visual impairment. CLEVR has a similar proposal to VQA and VizWiz, but was generated automatically by algorithms containing images with three basic shapes (spheres, cubes and cylinders) in two different sizes (small and large) and eight different colors, and includes questions and answers with the elements contained in the images. The combination of VQA, VizWiz and CLEVR gave origin to another dataset of questions and answers, annotated with the sensitive attribute of the visual conditions of the user who asked the question, which could be normal vision, visually impaired or robot.

The AFHQ \cite{jalal2021fairness} dataset is a dataset of animal faces consisting of 15,000 high-quality images at 512 × 512 resolution. It includes three domains of cat, dog and wildlife, each providing 5,000 images, it also contains three domains and several images of various breeds (larger than eight) for each domain. All images are aligned vertically and horizontally to have the eyes in the center. Low quality images were discarded. The work by \cite{jalal2021fairness} used only images of cats and dogs.

The Correlated and Imbalanced MNIST (CI-MNIST) \cite{reddy2021benchmarking} dataset is a variant of the MNIST dataset with additional artificial attributes for eligibility analysis. For an image, the label indicates eligibility or ineligibility, respectively, given that it is even or odd. The dataset varies the background colors as a protected or sensitive attribute, where blue denotes the non-privileged group and red denotes the privileged group. The dataset is designed to evaluate bias mitigation approaches in challenging situations and address different situations. The dataset has 50,000 images for the training set, 10,000 images for validation and testing with the eligible images representing 50 percent of each of these. Various background colors, colored boxes added at some top of the image of varying sizes were used to allow the impact of the colors, positions and sizes of the elements contained in the image to be analyzed.

Another alternative for the dataset is to create it synthetically \cite{gitiaux2019mdfa}, in which case it follows a normal distribution for the data. In it was created a binary sensitive attribute with Bernouilli distribution for its occurrence.

The use of the datasets presented can be seen in Section \ref{mitigation_technique} associated with mitigation techniques.

\subsection{Fairness metrics}

In the research \cite{paassen2019dynamic}, claims that machine learning models increasingly provide approaches to quantify bias and inequality in classification operations as a methodology for measuring bias and fairness. While many metrics have been developed, when it comes to long-term decisions, the models and scientific community have produced poor outcomes. Some existing metrics for measuring model bias are insufficient, either because they only evaluate the individual or the group, or because they are unable to predict a model's behavior over time. The authors offer the metric Demographic Parity as a solution, which when applied to a model ensures that the average classification of individuals in each group converges to the same point, achieving a balance between accuracy, bias, and fairness for the groups classified by the model.

In \cite{quadrianto2017recycling}, Demographic Parity, which assures that decisions are unconnected to sensitive attributes, was one of the metrics used to evaluate the model. Equalized Odds, to guarantee parity between positive and negative evaluations, and Equality of Opportunity, to ensure that individuals meet the same criteria and are treated equally. Each of these metrics assures that groups are treated fairly and that the model's quality does not deteriorate or become biased over time, as addressed in \cite{paassen2019dynamic}.

The metrics for assessing fairness should apply the same treatment to multiple groups, however if one of the metrics identifies bias, other metrics can charge that the model is fair.

Five metrics for assessing fairness were established from the review of the papers: Equalized Odds, Equality of Opportunity, Demographic Parity, Individual Differential Fairness, and MDFA.

As a basis for the fairness metrics, it is important to define true positive (TP), false positive (FP), true negative (TN), and false negative (FN). These values are obtained from the rights and wrongs of the model's prediction relative to the target or ground truth provided by the dataset. Positive values are defined as the positive class that the model should predict, as opposed to negative values. For example, if the model should predict whether an individual will reoffend, the positive class will be 1, which indicates that the individual will reoffend, and the negative class will be 0. Therefore, if the positive classes are correct, they will be computed in TP, while the errors will be computed in FP. On the other hand, hits for negative classes will be computed in TN and errors in FN. 

\begin{figure*}[h]
	\centering
		\includegraphics[scale=0.7]{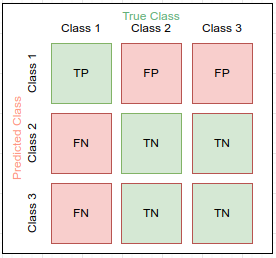}
	\caption{Confusion Matrix Multiclass}
	\label{FIG:confusion_matrix_multiclass}
\end{figure*}

For a multiclass problem there is no positive and negative class, just consider the values for each individual class, observe Figure \ref{FIG:confusion_matrix_multiclass}.

In the example, the scenario for calculating the values of Class 1 is illustrated, TP is the value of the correct prediction, consistent with the target. The TN are the sum of the classes that do not involve class 1, neither in the prediction, nor in the target. The PF are the sum of the classes falsely predicted as class 1, while the FN are the sum of the classes predicted as other classes that should have been predicted as class 1.

This process should be performed for all classes and the overall TP, FP, TN and FN of the model should be averaged over the individual values.

To understand the justice metrics, which use the TP, FP, TN and FN, the statistical metrics must also be defined as per Table \ref{tab:metrics_fairness}.
p{8em}p{35em}
\begin{table*}[h]
 \centering
\begin{tabular}{p{14em}p{15em}p{15em}}
\toprule
Statistical Metrics  & References & Equation \\ 
\midrule
Positive Predictive Value (PPV) & \cite{verma2018fairness} & $PPV = TP / (TP + FP)$ \\ \hline
False Discovery Rate (FDR) & \cite{verma2018fairness} & $FDR = FP / (TP + FP)$ \\ \hline
False Omission Rate (FOR) & \cite{verma2018fairness} & $FOR = FN / (TN + FN)$ \\ \hline
Negative Predictive Value (NPV)  & \cite{verma2018fairness} & $NPV = TN / (TN + FN)$ \\ \hline
True Positive Rate (TPR)  & \cite{verma2018fairness, quadrianto2017recycling, georgopoulos2021mitigating} & $TPR = TP / (TP + FN)$  \\ \hline
False Positive Rate (FPR) & \cite{verma2018fairness,quadrianto2017recycling, jain2019singular} & $FPR = FP / (FP + TN)$  \\ \hline
False Negative Rate (FNR) & \cite{verma2018fairness,jain2019singular} & $FNR = FN / (TP + FN)$ \\ \hline
True Negative Rate (TNR)  & \cite{verma2018fairness} & $TNR = TN / (FP + TN)$ \\
\bottomrule
\end{tabular}
\caption{Table of statistical metrics}
\label{tab:statistical_metrics}
\end{table*}

The objective of the metric Equalized Odds is to ensure that the probability that an individual in a positive class receives a good result and the probability that an individual in a negative class wrongly receives a positive result for the protected and unprotected groups are the same. That is, the TPR and FPR of the protected and unprotected groups must be the same. \cite{surveybias2019}.

\begin{equation}
 EO = \frac{1}{2} * \left ( \left | \frac{FP_{p}}{FP_{p}+TN_{p}} - \frac{FP_{u}}{FP_{u}+TN_{u}} \right | + \left | \frac{TP_{p}}{TP_{p}+FN_{p}}-\frac{TP_{u}}{TP_{u}+FN_{u}} \right |   \right )
 \label{eqn:EO}
\end{equation}

In contrast, the metric "Equality of Opportunity" must satisfy equal opportunity in a binary classifier (Z). As a result, the probability of an individual in a positive class receiving a good outcome must be the same for both protected and unprotected groups. That is, the TPR for both the protected and unprotected groups must be the same.\cite{surveybias2019}.

\begin{equation}
EOO = \frac{TP_{p}}{TP_{p}+FN_{p}}-\frac{TP_{u}}{TP_{u}+FN_{u}}
 \label{eqn:EOO}
\end{equation}

According to the justice metric Demographic Parity (DP), also known as Statistical Parity, the probability of an outcome being positive \cite{surveybias2019}. For this, the formula below should be applied.

\begin{equation}
DP = \frac{TP +FP}{N}
 \label{eqn:DP}
\end{equation}

The justice metric Disparate Impact (DI) compares the proportion of individuals who receive a favorable outcome for two groups, a protected group and an unprotected group. This measure must equal to 1 to be fair.

\begin{equation}
DI = \frac{\frac{T P_{p}+F P_{p}}{N_{p}}}{\frac{T P_{u}+F P_{u}}{N_{u}}}
 \label{eqn:DI}
\end{equation}

The K-Nearest Neighbors Consistency (KNNC) justice metric, on the other hand, is the only individual justice metric used by \cite{jang2021constructing}, it measures the similarity of sensitive attribute labels for similar instances \cite{bryant2019analyzing}.

\begin{equation}
KNNC = 1 - \frac{1}{n}\sum_{i=1}^n \left | {
\hat{y}}_i - \frac{1}{k} \sum_{j\in\mathcal{N}_{k}(x_i)} {\hat{y}}_j \right |
\end{equation}

Difference metrics were used as fairness metrics by \cite{jang2021constructing}, they are: Absolute Balanced Accuracy Difference (ABAD), Absolute Average Odds Difference (AAOD), Absolute Equal Opportunity Rate Difference (AEORD) and Statistical Parity Difference (SPD). The Differences metrics are calculated from the difference of the 'Disparity' metrics between two classes.

The ABAD is the difference in balanced accuracy in protected and unprotected groups, defined by Equation \ref{eqn:ABAD}.

\begin{equation}
ABAD = \left | \frac{1}{2} \left [ TPR_{p} + TNR_{p}  \right ] - \left [ TPR_{u} + TNR_{u}  \right ] \right |
 \label{eqn:ABAD}
\end{equation}

The AAOD is the absolute difference in TPR and FPR between different protected groups, defined by Equation \ref{eqn:AAOD}.

\begin{equation}
AAOD = \left | \frac{\left ( FPR_{u} + FNR_{p}  \right ) - \left ( TPR_{u} + TPR_{p}  \right )}{2} \right |
 \label{eqn:AAOD}
\end{equation}

AEORD is the difference in recall scores (TPR) between the protected and unprotected groups. A value of 0 indicates equality of opportunity, defined by Equation \ref{eqn:AEORD}.

\begin{equation}
AEORD = \left | TPR_{p} - TPR_{u} \right |
\label{eqn:AEORD}
\end{equation}

Finally, SPD is the difference in SD between a protected and an unprotected group, defined by Equation \ref{eqn:SPD}.

\begin{equation}
SPD = \frac{TP_{p} +FP_{p}}{N_{p}}-\frac{TP_{u}+FP_{u}}{N_{u}}
 \label{eqn:SPD}
\end{equation}

In addition to fairness metrics, some works use classification metrics such as accuracy, precision, recall and F1-score \cite{das2019dataset} as criteria for identifying bias. In addition to fairness metrics, some works use classification metrics such as accuracy, precision, recall and F1-score \cite{das2019dataset} as criteria for identifying bias. Measures of bias linked to the accuracy of model predictions are designed to check for unexpected differences in accuracy between groups. A less accurate prediction for one group compared to another contains systematic error, which disproportionately affects one group over the other \cite{booth2021integrating}

Accuracy is the ratio between the number of true negatives and true positives to the total number of observations. Precision is the proportion of correct positive identifications. Recall is the proportion of true positives correctly identified. The F1-score is the weighted average of Precision and Recall. The formulas for each can be seen in Equations \ref{eqn:acc}, \ref{eqn:recall}, \ref{eqn:f1}, \ref{eqn:f1}

\begin{equation}
accuracy = \frac{TN + TP}{TN+TP+FN+FP}
 \label{eqn:acc}
\end{equation}

\begin{equation}
precision = \frac{TP}{TP+FP}
 \label{eqn:precision}
\end{equation}

\begin{equation}
recall = \frac{TP}{TP+FN}
 \label{eqn:recall}
\end{equation}

\begin{equation}
\textrm{F1-score} = \frac{2 * \left (recall * precision  \right )}{\left (recall * precision  \right )}
 \label{eqn:f1}
\end{equation}

Other cases used the number of positive (NIP) and negative (NIN) instances as the criteria for fairness metrics, as well as the base rate (BR) also known as prior probabilities are the unconditional probabilities, it is a probability with respect to all samples (N) \cite{bryant2019analyzing}. The formulas for each can be seen in Equations \ref{eqn:nip}, \ref{eqn:nip}, \ref{eqn:br}

\begin{equation}
NIP = TP + FP
 \label{eqn:nip}
\end{equation}

\begin{equation}
NIN = TN + FN
 \label{eqn:nin}
\end{equation}

\begin{equation}
BR = NIP / N
 \label{eqn:br}
\end{equation}

All reported fairness metrics can be seen in Table \ref{tab:metrics_fairness}.

\begin{table*}[h]
\setlength{\tabcolsep}{18pt}
\renewcommand{\arraystretch}{0.4}
 \centering
\begin{tabular}{p{7em}p{30em}}
\toprule
Metric Fairness  & References \\ \midrule
EO & \cite{surveybias2019, georgopoulos2021mitigating, reddy2021benchmarking, radovanovic2020enforcing, du2021fairness}  \\ \hline
EOO & \cite{surveybias2019, reddy2021benchmarking, amend2021improving, radovanovic2020enforcing, li2022fair} \\ \hline
DP & \cite{surveybias2019,jalal2021fairness, reddy2021benchmarking, amend2021improving, du2021fairness, li2022fair} \\ \hline
DI & \cite{bryant2019analyzing} \\ \hline
KNNC & \cite{bryant2019analyzing} \\ \hline

ABAD & \cite{jang2021constructing} \\ \hline
AAOD & \cite{jang2021constructing} \\ \hline
AEORD & \cite{jang2021constructing} \\ \hline
SPD & \cite{jang2021constructing} \\ \hline

accuracy & \cite{das2019dataset} \\ \hline
precision  & \cite{das2019dataset} \\ \hline
recall & \cite{das2019dataset} \\ \hline
F1-score  & \cite{das2019dataset} \\ \hline

NIP  & \cite{bryant2019analyzing} \\ \hline
NIN & \cite{bryant2019analyzing} \\ \hline
BR  & \cite{bryant2019analyzing} \\
\bottomrule

\end{tabular}
\caption{Metrics used as fairness criteria}
\label{tab:metrics_fairness}
\end{table*}

\subsection{Techniques for Bias Analysis}

For bias analysis and identification, a methodology was proposed by \cite{fontanamonitoring} for models trained with federated learning with the HOLDA architecture, checking the influence of biased individuals on unbiased individuals. Whenever a user updates its internal state by replacing the previous best model, when that model has a better generalization performance on the local validation data, the system evaluates the fairness of that new model. They performed an experiment training an ANN with 200 neurons in the hidden layer. The sensitive attribute used was "Gender". They concluded that local models trained with unbiased customers have little influence on the model, while biased customers impact the model unfairness. In this way the biased customers end up influencing the unbiased ones, but local models trained only with an unbiased customer tend to be slightly unfair. The dataset used was Adult. The fairness metrics used were DP, EO and EOO.

Also aiming to identify unfairness and promote explainability of model decisions, the technique suggested by \cite{seymour2018detecting} includes a model that combines white-box and black-box features for local and global explanations, respectively. Local explanation involves determining which features contributed the most to the classification of a given data sample, which can be achieved with a visualization tool or algorithm that can simulate and explain the decisions of the original model. In terms of overall explanation, the model decisions perform comparisons with the classifications obtained by each group, using decile risk scores to demonstrate whether there is bias in the model. The experiments were performed with the COMPAS dataset.

\subsection{Mitigation techniques and models} \label{mitigation_technique}

As noted earlier, bias and unfairness mitigation techniques can be of the types: pre-processing, in-processing, and post-processing. According to \cite{di2022recommender} pre-processing mitigation techniques focus on rebalancing the data. In-processing mitigation, on the other hand, focuses on the model and its regularization with a bias correction term in the loss function or implicit in the model as with adversarial networks, where the model predicts the sensitive attribute.

The preprocessing mitigation technique aims to alter the dataset in a way that positively impacts the fairness metrics, as can be seen in \cite{yang2020fairness}'s study, where FairDAGs library is proposed as an acyclic graph generator that describes the data flow during preprocessing. Its purpose is to identify and mitigate bias in the distribution and distortions that may arise with protected groups, while allowing direct observation of changes in the dataset. The four types of treatment are: bias by filtering the data, standardizing missing values, changes in the proportion of the dataset after replacement of NaN values, and, for natural language processing (NLP) systems, filtering out extraneous names or words that the computer may not recognize. The results showed that DAG was able to identify and represent differences in the data that occurred during preprocessing, as well as correct imbalances in the datasets examined.

In the work of \cite{bryant2019analyzing} the preprocessing has a different purpose, as it aims to remove sensitive data from the model for a banking system, ensuring the removal of customer data after the output without affecting the ML model. The goal is the generation of synthetic data from the representation of the original data in order to preserve privacy while maintaining the usefulness of that original data. The synthetic data is generated by the Trusted Model Executor (TME) which is an AIF360 tool. At the end, the bias in the synthetic dataset was evaluated by comparing it with the original datasets in order to validate the TME.

Also using AIF360 to perform preprocessing operations, the study by \cite{paviglianiti2020vital} addresses that smartwatches distinguish between men and women in the identification of cardiovascular problems, evaluating more characteristics of the former group than the latter. In view of the above there should be a correction to fit the needs of both genders the removal of sensitive data, with the rebalancing of the dataset distribution and processing operations. It also adjusts non-representative data for accurate assessment of user health. The mitigation technique in preprocessing used was Reweighting. At the end the Vital-ECG was developed, a watch-like device that detects heart rate, blood pressure, skin temperature and other body variables without distinction of gender and with superior predictions.

Still in the area of data generation, the work \cite{jang2021constructing} generates a new dataset that has no disparity of distribution, quality or noise, ensuring that all classes are treated equally. To do this it used the VAE-GAN architecture which, although it showed great improvements in model impartiality, the use of synthetic data during training limited its ability to generalize real data, reducing accuracy and precision. To minimize the trade-off, the model trained with artificial data used transfer learning techniques to perform an adjustment of the weights with real data.

In \cite{georgopoulos2021mitigating} work, in the area of computer vision, highlights that face recognition and analysis models generally exhibit demographic biases, even in models where accuracy is high. The reason is usually due to datasets with under-represented categories, whether for identifying identity, gender, or expressions of the human face. Biases can be in relation to age, gender, and skin tone. Therefore, a bias mitigation technique was proposed with a dataset of facial images, where to increase demographic diversity, a style transfer approach using Generative Adversarial Networks (GANs) was used to create additional images by transferring multiple demographic attributes to each image in a biased set. The literature review study \cite{jinyin2021fairness} highlights pre-processing techniques to mitigate data bias, such as Synthetic Minority Over-sampling Technique (SMOTE) and uses Data Augmentation, as well as \cite{georgopoulos2021mitigating}. In the end it defines open questions on the topic such as the fact that metrics can be conflicting, indicating a model that is fair in one metric and unfair in another.

Also in the area of computer vision, \cite{jalal2021fairness} presents several intuitive notions of group fairness, applied to image enhancement problems. Due to the uncertainty in defining the clusters, since, for the author, there are no ground truth identities in the clusters, and the sensitive attributes are not well defined.

Some conclusions about the impacts of fairness metrics on the pre-processing mitigation process could be obtained in \cite{jalal2021fairness}. It states that the metric \textit{demographic parity} is strongly dependent on clusters, which is problematic for generating images of people in the data augmentation process, because the classes of the sensitive attribute 'race' are ill-defined. In CPR, implemented using \textit{Langevin dynamics}, this phenomenon does not occur, and it can be seen in the results obtained that, for any choice of protected clusters, the expected properties are displayed.

The fairness metrics identified in the papers that addressed preprocessing are in Table \ref{tab:metrics_preprocess}, as are the datasets in Table \ref{tab:datasets_preprocess}.

\begin{table}[H]
 \centering
\begin{tabular}{cc}
\toprule
Fairness Metrics  & References \\ \midrule
FPR                  & \cite{yang2020fairness} \\ \hline
FNR                  & \cite{yang2020fairness} \\ \hline
Demographic Parity                 & \cite{jalal2021fairness} \\ \hline
TPR                 & \cite{georgopoulos2021mitigating}  \\ \hline
Equalized Odds                 & \cite{georgopoulos2021mitigating}  \\ \hline
Absolute Balanced accuracy difference                 & \cite{jang2021constructing} \\ \hline
Absolute average odds difference                 & \cite{jang2021constructing} \\ \hline
Absolute equal opportunity rate difference                 & \cite{jang2021constructing} \\ \hline
Statistical Parity Difference                 & \cite{bryant2019analyzing} \\ \hline
Disparate Impact                 & \cite{bryant2019analyzing} \\ \hline
K-Nearest Neighbors Consistency                 & \cite{bryant2019analyzing} \\ \hline
Number of Positive Instances                 & \cite{bryant2019analyzing} \\ \hline
Number of Negative Instances                 & \cite{bryant2019analyzing} \\ \hline
Base Rate                 & \cite{bryant2019analyzing} \\ \bottomrule

\end{tabular}
\caption{Fairness metrics used in preprocessing techniques}
\label{tab:metrics_preprocess}
\end{table}

\begin{table}[h]
 \centering
\begin{tabular}{cc}
\toprule
Datasets  & References \\ 
\midrule
Heart                 & \cite{jinyin2021fairness} \\ \hline
Adult                 & \cite{jinyin2021fairness, yang2020fairness, jang2021constructing, bryant2019analyzing} \\ \hline
Bank marketing dataset                 & \cite{jinyin2021fairness, bryant2019analyzing} \\ \hline
Boston house price dataset                 & \cite{jinyin2021fairness} \\ \hline
COMPAS dataset                 & \cite{jinyin2021fairness, yang2020fairness, jang2021constructing}  \\ \hline
German credit dataset                 & \cite{jinyin2021fairness, jang2021constructing} \\ \hline
Medical Expenditure Panel Survey (MEPS)                 & \cite{jinyin2021fairness, jang2021constructing} \\ \hline
FlickrFaces                & \cite{jalal2021fairness} \\ \hline
AFHQ Cats and Dogs                 & \cite{jalal2021fairness} \\ \hline
LFW                 & \cite{georgopoulos2021mitigating} \\ \hline
CelebA                & \cite{georgopoulos2021mitigating} \\ \hline
MOPRH                 & \cite{georgopoulos2021mitigating} \\ \hline
MIMIC II                 & \cite{paviglianiti2020vital} \\ \bottomrule
\end{tabular}
\caption{Datasets used in preprocessing techniques}
\label{tab:datasets_preprocess}
\end{table}

The in-processing mitigation technique was identified in a larger amount of papers, and can be observed in \cite{martinez2021using, radovanovic2020enforcing, jain2019singular, reddy2021benchmarking, ashokan2021fairness, adel2019one, amend2021improving, quadrianto2017recycling, cerrato2020constraining}

An in-processing solution was proposed by \cite{martinez2021using} in the holistic and often subjective methods that may contain biases in the student selection process in schools. From this perspective, learning algorithms capable of admitting a diverse student population were developed, even for groups with historical disadvantages. The study examined the impact of characteristics such as income, color, and gender on student admission rates.

The work \cite{radovanovic2020enforcing} also presents an in-process mitigation solution for group bias. It used the logistic regression technique to develop the model.  The solution used was Pareto Optimal, which aims to ensure a better accuracy loss function while keeping the fairness metrics at the threshold set at 80\%. The author states that the in-processing solution, where the algorithm is adjusted during learning, would be a natural solution, because the pre-processing algorithms would be altering the original data, hurting ethical norms, however it is possible to work with data balance, without altering the users' data.

Another em-processing mitigation model was proposed by \cite{jain2019singular}, with a new classification approach for datasets based on the sensitive attribute 'race', with the aim of increasing prediction accuracy and reducing racial bias in crime recidivism. The recidivism prediction models, were evaluated by the type of crime, including 'violent crimes', 'property', 'drug' and 'other'. For the 'all crimes', 'Caucasian data set', and 'African American data set' groups, the results still contained bias, although lower than the baseline data. The ratios obtained were 41:59, 34:66, and 46:54.

The \cite{reddy2021benchmarking} study focuses on bias mitigation in deep learning models for classification. The authors point to the need for a systematic analysis of different bias mitigation techniques in-processing with MLP and CNN. Using a dataset that allows the creation of different bias sets, they performed an analysis of the mitigation models recently proposed in the literature. Then they showed the correlation between eligibility and sensitive attributes, the possible presence of bias even without sensitive attributes, and the importance of the initial choice of architecture for model performance.

Whereas \cite{ashokan2021fairness} presents a focus on the ways in which bias can occur in recommender systems, while addressing the lack of systematic mapping to address unfairness and bias in the current literature. In the experiments, sources of unfairness that can occur in recommendation tasks were mapped, while evaluating whether existing bias mitigation approaches successfully improve different types of fairness metrics. It also presents a mitigation strategy in which the algorithm learns the difference between predicted and observed ratings in subgroups, identifying which is biased and correcting the prediction. The results show that fairness increased in most use cases, but performance for MSE and MAE vary in each case.

The works of \cite{martinez2021using, radovanovic2020enforcing, jain2019singular, reddy2021benchmarking, ashokan2021fairness, zheng2021cost} have in common the fact that their models were trained in order to mitigate bias from only adjusting the weights of their proposed models. In the work of \cite{du2021fairness} there is already an attempt to mitigate the bias by neutralizing the sensitive attribute in the model. It has been shown to be possible to make a classification model fairer by removing bias only in its output layer, in a process that occurs during its construction. To this end, a technique was developed where training samples with different sensitive attributes are neutralized, causing the model's dependence on sensitive attributes to be reduced. The main advantage demonstrated by the method is the small loss of accuracy in exchange for improved fairness metrics, without requiring access to the sensitive attributes in the database. In addition, the authors argue that it is possible to increase the quality of the technique by combining it with others, for example by using a fairer basis than the one used in the experiments. 

In the solution proposed by \cite{zheng2021cost} the classification detects the item with the highest probability of belonging to the 'target' class of the model; however, there are cases where numerous items have very close probabilities and bias the model, causing an error to propagate across multiple levels. To avoid this, there is a need for a threshold with a minimum degree for the data to be classified and triggers a recalculation of the maximum node probability. The sensitive cost then performs its own probability calculation on the data with the highest degree of membership. These calculations avoid bias caused by using a single probability or over-optimal adjustment caused by using data with no prior context. Hierarchical Precision and Hierarchical Precision, which evaluates the relationship between all descendants of the class and includes Hierarchical F1, Hierarchical Recall, and Hierarchical Precision, were used as metrics. The threshold is adaptive, without requiring user parameters, since metrics exist throughout the classification. Even with fewer samples, it produced results that were superior to the state of the art.

The other works the neutralization of sensitive attributes in an attempt to mitigate model bias is more direct as can be seen in \cite{adel2019one, amend2021improving, quadrianto2017recycling, cerrato2020constraining} by identifying it beforehand, similarly the investigation of \cite{chiappa2018causal} which addresses a new perspective on the concept of justice by determining whether an attribute is sensitive by evaluating it in a Causal Bayesian Networks model. This model examines the direct effects of one characteristic on another and determines whether a sensitive attribute 'A' influences the output 'Y' of a model, producing correlation plots that strive to understand whether or not decisions made were made fairly.

In \cite{adel2019one} a pre-existing biased model must be updated to become fair, minimizing unfairness without causing abrupt structural changes. The study uses an adversarial learning technique with the distinction that the generating model is the original network, however the adversarial model comprises an extra hidden layer, rather than a second model, in order to predict which sensitive attribute influenced the generator's decision. The main element of this competition model is that if the discriminator finds the sensitive attribute that influenced the decision the most, it demonstrates dependence on the generator model, suggesting bias. The generator moves away from the sensitive attributes and performs a classification that does not depend on them, eventually lowering the discriminator's hit rate until it completely loses its predictive ability. The network architecture has three parts: adding an adversarial layer on top of the network, balancing the distribution of classes across the minisets, and adapting sensitive attributes until they are no longer present.

The technique was developed for classification tasks, but can be used for any neural network with biases starting with sensitive attributes \cite{adel2019one}, achieved better results compared to the state of the art with the metrics addressed.

In the same way as \cite{adel2019one}, \cite{amend2021improving} also uses adversarial network for sensitive attribute identification and examines metrics and combinations of techniques for bias mitigation. The study was conducted using basic ANN models and a Split model, which forms the basic model by permuting attribute classes as training criteria in order to identify which one is sensitive. Another model based on the Classifier-Adversarial Network (CAN) architecture, in which the adversarial network predicts the sensitive attribute based on the output of the basic model. Finally, there is the CAN with Embedding (CANE) architecture, which takes as input the output of the basic model as well as the weights created in the penultimate layer. They demonstrated that the models from the Basic RNA architecture can improve accuracy, but not bias. Meanwhile, the models of the CAN and CANE architectures improved accuracy and reduce bias, with CANE being better than CAN.

Still involving adversarial network, in \cite{li2022fair} the Adversarial Fairness Local Outlier Factor (AFLOF) method is proposed for outlier detection, combining adversarial algorithms with the Local Outlier Factor (LOF) algorithm, which returns a value indicating whether an instance is an outlier, aiming to achieve a fairer and more assertive result than LOF and FairLOF. It works with the sensitive attributes "Gender", "Age" and "Race". It also uses the AUC-ROC score to measure outlier detection. It results in a fairer and more assertive performance for outlier detection than the previous methods cited, thus achieving a breakthrough in the study of fairness. 
The work of \cite{grari2019fair}, on the other hand, argues that research on fairness and bias in machine learning focuses only on neural networks, with few publications for other classification techniques. As a result, the author investigated Adversarial Gradient Tree Boosting to rank data and noted that while the adversary progressively loses the reference of the sensitive attribute that led to that prediction.

Another contribution is the adversarial learning method for generic classifiers, such as decision trees in \cite{grari2019fair}. Comparing numerous state-of-the-art models with the one provided in the paper, which covers two justice metrics. They used varied decision trees in the model given that they make rankings, which are then sent through a weighted average to an adversary, who predicts which sensitive attribute was significant to the final decision. While the adversary is able to detect the sensitive attribute, a gradient propagation occurs, updating the weights in the decision trees and trying to prevent the sensitive attribute from having a direct impact on the ranking.

The \cite{grari2019fair} model called FAGTB performed well on accuracy and fairness metrics for the COMPAS, Adult, Bank, and Default datasets, outperforming other state-of-the-art models on several of them and considerably outperforming the network adversary. The study leaves certain questions unanswered for future research, such as an adversary using Deep Neural Decision Forests. If this method were used to retrieve the gradient, theoretically, the transparency of the model for the algorithm's decision would be apparent because it consists only of trees. As a final caveat, they acknowledge that the algorithm handles distinct groups well, but the EO and DP fairness metrics do not measure bias between individuals, and is an aspect for improvement.

Following varied work with adversarial learning, the model proposed by \cite{quadrianto2017recycling} called Privileged Information is a technique that trains the model with all the features of the original dataset, including sensitive attributes, and then tests it without these attributes. The model is an in-processing type adjusted with the goal of mitigating unfairness and independent of sensitive attributes, while maintaining its ability to produce accurate predictions, thus respecting the protected information for decision making. Note that in this case, the model fully fits the dataset in an attempt to mitigate bias. The author emphasizes the strength of his model in identifying the best predictor relative to other state-of-the-art work, having the sensitive attributes as optional, and still using Privileged Information.

Whereas \cite{cerrato2020constraining} avoids model bias by using only data with minimal or, if possible, no sensitive attributes. By applying a noise conditioning operation to the data provided in the model, inducing the model to ignore sensitive attributes, reducing bias. The goal of the model is to create as accurate a representation as possible in the prediction, with fairness. The models used the techniques of logistic regression and Random Forest.

The justice metrics identified in the papers that addressed in-processing are in Table \ref{tab:metrics_inprocess}, as are the datasets in Table \ref{tab:datasets_inprocess}.

\begin{table*}[h]
 \centering
\begin{tabular}{p{10em}p{38em}}
\toprule
Fairness Metric  & References \\ \midrule
Demographic Parity                  & \cite{reddy2021benchmarking, amend2021improving, du2021fairness, li2022fair} \\ \hline
Equality of Opportunity                 & \cite{reddy2021benchmarking, amend2021improving, radovanovic2020enforcing, li2022fair} \\ \hline
Equalized Odds                & \cite{reddy2021benchmarking, radovanovic2020enforcing, du2021fairness} \\ \hline
accuracy                 & \cite{reddy2021benchmarking, quadrianto2017recycling, amend2021improving, jain2019singular} \\ \hline
Disparate Impact                 & \cite{adel2019one}  \\ \hline

TPR                 & \cite{quadrianto2017recycling} \\ \hline
FPR                 & \cite{quadrianto2017recycling, jain2019singular, adel2019one} \\ \hline
FNR                 & \cite{jain2019singular, adel2019one} \\ 
\bottomrule
\end{tabular}
\caption{Fairness metrics used in em-processing techniques}
\label{tab:metrics_inprocess}
\end{table*}


\begin{table*}[H]
 \centering
\begin{tabular}{p{10em}p{38em}}
\toprule
Datasets  & References \\ 
\midrule
CI-MNIST                 & \cite{reddy2021benchmarking} \\ \hline
Adult                 & \cite{reddy2021benchmarking, adel2019one, quadrianto2017recycling, amend2021improving, cerrato2020constraining, radovanovic2020enforcing, du2021fairness, li2022fair}  \\ \hline
COMPAS                 & \cite{adel2019one, quadrianto2017recycling, cerrato2020constraining, radovanovic2020enforcing} \\ \hline
German credit                 & \cite{cerrato2020constraining} \\ \hline
Bank Marketing                 & \cite{cerrato2020constraining}  \\ \hline
FDOC                 & \cite{jain2019singular} \\ \hline
FDLE                & \cite{jain2019singular} \\ \hline
MovieLens 1M                 & \cite{ashokan2021fairness}  \\ \hline
MEPS                & \cite{du2021fairness} \\ \hline
CelebA                 & \cite{du2021fairness} \\ \hline
Weight                 & \cite{li2022fair} \\ \hline
Drug                & \cite{li2022fair} \\ \hline
Crime                 & \cite{li2022fair} \\ \hline
Student                 & \cite{li2022fair} \\ \hline
Credit                 & \cite{li2022fair} \\
\bottomrule
\end{tabular}
\caption{Datasets used in the in-processing techniques}
\label{tab:datasets_inprocess}
\end{table*}

Mitigation solutions for post-processing were also found, as in \cite{gitiaux2019mdfa, das2019dataset}. In \cite{gitiaux2019mdfa} it proposes a solution for an already formed model, seeking to identify whether certain groups receive discriminatory treatment due to their sensitive attributes. With the identification of discrimination for a group, it is verified whether the sensitive attributes are impacting the model, even if indirectly. The model has a neural network with four fully connected layers of 8 neurons, expressing the weights as a function of the features in order to minimize the maximum average discrepancy function between the sensitive attribute classes promoting unfairness mitigation. He applied his mitigation model to a Logistic Classification model. The work allows black-box type models to be mitigated for unfairness, but also by understanding the assigned treatment.

In the \cite{das2019dataset} study it uses the identification of biases in models developed to recognize the user, where the user can be a human with normal vision, a blind person, or a robot. The identification takes place when answering a question, so NLP is applied. Its bias can be seen in the most frequently asked question "what is this object?", as well as the low image quality compared to the others. Initially, annotations were assigned to the content of the images such as "boy", "package", "grass", "airplane", and "sky". Random Forest, K-Nearest Neighbors, Nave Bayes, and Logistic Regression techniques were used to develop the models. Logistic Regression produced the best results, with 99\% on all metrics. The authors found that the algorithms readily recognized the bias in each dataset and provided a means of tracing the origin of the questions and images.

The justice metrics identified in the papers that addressed post-processing are in Table \ref{tab:metrics_postprocess}, as are the datasets in Table \ref{tab:datasets_postprocess}.

\begin{table*}[H]
 \centering
\begin{tabular}{p{10em}p{38em}}
\toprule
Fairness Metric  & References \\ \midrule
disparate impact                 & \cite{gitiaux2019mdfa} \\ \hline
precision                & \cite{das2019dataset} \\ \hline
recall                 & \cite{das2019dataset} \\ \hline
accuracy                 & \cite{das2019dataset}  \\ \hline
F1-score                & \cite{das2019dataset}  \\ 
\bottomrule

\end{tabular}
\caption{Fairness metrics used in post-processing techniques}
\label{tab:metrics_postprocess}
\end{table*}


\begin{table*}[h]
 \centering
\begin{tabular}{p{12em}p{36em}}
\toprule
Datasets  & References \\
\midrule
Sintetic (normal distribution)                  & \cite{gitiaux2019mdfa} \\ \hline
COMPAS                 & \cite{gitiaux2019mdfa}  \\ \hline
VQA                 & \cite{das2019dataset} \\ \hline
VizWiz                 & \cite{das2019dataset} \\ \hline
CLEVR                 & \cite{das2019dataset}  \\
\bottomrule
\end{tabular}
\caption{Datasets used in post-processing techniques}
\label{tab:datasets_postprocess}
\end{table*}

\section{Discussion}

All 45 studies examined addressed comparable techniques, case studies, datasets, metrics, and applications. 

Adult datasets and COMPAS were used to address the most frequently reported bias identification and injustice mitigation.

In \cite{jinyin2021fairness}'s work investigated the sources and implications of various types of bias, either in the datasets or in the model. The study investigates bias, offering methods for eliminating it, as well as constructing groups and subgroups that help understand the problem, and discusses general categories such as temporal, spatial, behavioral, posterior, transcendental, and group bias. Specific cases, such as the Simpsons paradox or social behavior bias, are grouped within these categories.

The forms of bias observed by \cite{jinyin2021fairness} are categorized as follows: dataset bias, model bias, and emergent bias, or pre-processing, in-processing, and post-processing, as previously described. In order to go deeper into these categories, the study \cite{jinyin2021fairness} splits them into eight broad and 18 particular categories, as well as providing metrics and strategies for resolving each of them.

A frequent concern about the individual-group interaction is that few ML models handle it. According to \cite{mitchell2021algorithmic}, if a model is biased in rejecting loans to black males, for example, it will increase its database with rejections for this group, reinforcing the bias and initiating a vicious spiral that will reassert itself with each loan denial.

The work \cite{paassen2019dynamic} focuses on the topic of vicious loops in machine learning, claiming that models may be free of bias in the present but may be biased in the future. To overcome this, he suggests that the model fulfill the Demographic Parity metric, which ensures that the classification of varied groups is constantly converging and that no group is disadvantaged over time.

Except for \cite{gitiaux2019mdfa} and \cite{seymour2018detecting}, the model proposals were primarily white-box classification. The former proposes a model for bias elimination using Multi-Differential Fairness by integrating in-processing and post-processing, whereas the latter proposes that the focus of algorithm transparency should be on the output rather than the whole decision-making process of the algorithm.

According to the works reviewed, sensitive attributes are defined as elements that should not directly affect the prediction of a model, such as color, race, sex, nationality, religion, and sexual preference, among others. According to US laws such as the \textit{Fair Housing Act} (FHA) and the \textit{Equal Credit Opportunity Act} (ECOA) \cite{CreditOpportunityAct}, sensitive attributes should never favor, harm, or alter the outcome of individuals and groups in decision-making processes such as hiring or a court sanction.
There is also the fact that all techniques and tools confirm the importance of sensitive attributes in mitigating biases, because for the identification of bias there is the need for the indication of a sensitive attribute, and the mitigation of bias will be based on this identification, remembering that the identification is done through a justice metric.

As for the datasets, 25 datasets were identified, most of them with sensitive attributes such as demographic data, and the ones that did not have any were for studies in the area of image enhancement, when not associated with face recognition. The datasets address aspects related to criminality, the selection and approval process of individuals, financial issues of bank credit, product pricing, health and medical diagnosis, face recognition and image enhancement, and synthetic datasets.

About the metrics of justice, the most used are EO, EOO and DP, as observed in Table \ref{tab:metrics_fairness}. Highlighting the importance of statistical metrics, difference metrics, and classification metrics, as several papers have used them as criteria for fairness.

Among the bias mitigation and identification tools: FairLearn and AIF360, weren't  used in any practical studies. The topic of identifying bias in the data and the model was also addressed, with the Aequitas tool being the most frequently mentioned.

As for the mitigation techniques, pre-processing techniques for rebalancing the data were addressed; in the in-processing techniques, such as regularizing the model, addressing levels of elimination of the sensitive attribute, with some possible approaches, such as being identified before training the model, during model training, not using it in training or disregarding it completely, training with all attributes so that the model can adjust itself through a loss function. The post-processing techniques, on the other hand, aim to discover which sensitive attribute had an impact on the model result, rebalancing the prediction.

The most common justice metrics such as EO, EOO and PD are covered with the FairLearn, AIF360, Aequitas and Responsible AI tools, with the exception of EOO not covered by FairLearn.

In \cite{di2022recommender}'s work highlights some research gaps such as the wide varieties of justice metrics as a factor hindering which one best fits each case, lacking a comprehensive formal and comparative study of the strengths and limitations of each of the metrics. It also highlights that a formal study of the techniques with the strengths and limitations of each is lacking. It also addresses the need for state-of-the-art recommendation system techniques. It highlights that there is still an absence of studies on the economic and social consequences of biases in high-risk systems. The work of \cite{kordzadeh2022algorithmic} attempts to elucidate some of these gaps, from the point of view of organizations and individuals, but without addressing the technical aspects of such solutions, when it highlights the importance of the socio-technical nature of biases in algorithms, the need to understand the social processes and contexts impacted by the use of biased information and algorithmic technologies.

Finally, all studies have addressed the algorithm's transparency, or the capacity to explain the decision-making process that caused the model to classify a certain individual or group the way it did. This method must fundamentally explain either the local decision, which includes the classification of a single individual, or the global decision, which verifies the whole algorithm process. The relevance of transparency is to make it explicit to a customer, company, or court that the model does not consider sensitive attributes and does not discriminate against a specific group, just as it becomes possible to attribute responsibility to the model's developers if the model is biased.

\section{Final considerations}

The objective of this study was to examine the latest existing knowledge on bias and unfairness in machine learning (ML) models with the RSL methodology and a bibliometric analysis. Thus, the paper was to answer questions Q1 and Q2 in the \ref{sec:Literature_Search} section.

To answer question Q1, the findings demonstrate that there is a focus on bias and unfairness identification methods for ML technologies, with well-defined metrics in the literature, such as fairness metrics, featured in tools, datasets, and bias mitigation techniques. This diversity ends up not defining the most appropriate approach for each context given that different solutions can be observed for the same problem, leading to a lack of definition about which one would be the most appropriate, without a generic solution for the identification and mitigation of biases. The vagueness raised in Q1's answer opens up aspects to be considered in Q2's answer.

To answer question Q2, where the existing opportunities should be highlighted, there is very limited support for black-box models, which contrasts with the abundance of information for white-box models. The need for transparency and explainability of ML algorithms, as well as the defining and preservation of sensitive attributes was also emphasized, with the selected datasets acting as a basis for research addressing the identification and mitigation of bias and unfairness.

As opportunities for future work, we conclude that more research is needed to identify the techniques and metrics that should be employed in each particular case in order to standardize and ensure fairness in machine learning models. For a definition on which metric should be used for each use case, more specific studies should be conducted under different architectures and sensitive attributes. This analysis would allow the context to define the most appropriate metric to identify bias in protected groups, and whether the sensitive attribute can be a relevant element in defining the fairness metric for a given context. It was observed that, in a given dataset, the metrics do not present uniform results, pointing to different categories of bias and their context-related particularities.

\bibliographystyle{unsrt}  
\bibliography{references}

\end{document}